\documentclass{article}

\usepackage{microtype}
\usepackage{graphicx}
\usepackage{subcaption}
\usepackage{booktabs} 

\usepackage{hyperref}
\usepackage{array}

\PassOptionsToPackage{sort&compress}{natbib}
\usepackage[preprint]{icml2026}
\setcitestyle{numbers,square,citesep={,},sort&compress}

\usepackage{amsmath}
\usepackage{amssymb}
\usepackage{mathtools}
\usepackage{amsthm}
\usepackage[normalem]{ulem}

\usepackage[capitalize,noabbrev]{cleveref}

\theoremstyle{plain}

\theoremstyle{definition}

\theoremstyle{remark}

\usepackage[textsize=tiny]{todonotes}

\usepackage{enumitem}
\usepackage{xcolor}
\usepackage[most]{tcolorbox}
\usepackage[acronym]{glossaries}
\usepackage{xspace}

\makeglossaries

\crefname{section}{\S}{\S\S}

\newacronym{llm}{LLMs}{Large Language Models}
\newacronym{dpr}{DP}{Differential Privacy}
\newacronym{ci}{CI}{Contextual Integrity}
\newacronym{he}{HE}{Homomorphic Encryption}
\newacronym{mpc}{MPC}{Secure Multi-Party Computation}
\newacronym{nlp}{NLP}{Natural Language Processing}
\newacronym{tee}{TEE}{Trusted Execution Environments}
\newacronym{pii}{PII}{Personally Identifiable Information}
\newacronym{ner}{NER}{Named Entity Recognition}

\newcommand{\llm}{\gls{llm}\xspace}
\newcommand{\dpr}{\gls{dpr}\xspace}
\newcommand{\ci}{\gls{ci}\xspace}
\newcommand{\he}{\gls{he}\xspace}
\newcommand{\mpc}{\gls{mpc}\xspace}
\newcommand{\nlp}{\gls{nlp}\xspace}
\newcommand{\tee}{\gls{tee}\xspace}
\newcommand{\pii}{\gls{pii}\xspace}
\newcommand{\ner}{\gls{ner}\xspace}

\icmltitlerunning{Privacy in Context Is Being Overlooked in the Age of General-Purpose LLMs}

\begin{document}

\twocolumn[
  \icmltitle{Privacy in Context Is Being Overlooked in the Age of General-Purpose LLMs}



  \icmlsetsymbol{equal}{*}

  \begin{icmlauthorlist}
    \icmlauthor{Prakhar Ganesh}{mcgill,mila,equal}
    \icmlauthor{Yash More}{mcgill,mila,equal}
    \icmlauthor{Marco Romanelli}{hofstra}
    \icmlauthor{Ferdinando Fioretto}{virginia}
    \icmlauthor{Golnoosh Farnadi}{mcgill,mila}
  \end{icmlauthorlist}

  \icmlaffiliation{mcgill}{McGill University}
  \icmlaffiliation{mila}{Mila - Quebec AI Institute}
  \icmlaffiliation{hofstra}{Hofstra University}
  \icmlaffiliation{virginia}{University of Virginia}

  \icmlcorrespondingauthor{Prakhar Ganesh}{prakhar.ganesh@mila.quebec}


  \vskip 0.3in
]



\printAffiliationsAndNotice{\icmlEqualContribution}

\begin{abstract}
As \llm continue to grow in scope and capability, privacy concerns are on the rise. Since users increasingly interact with a \textit{single general-purpose system} across many distinct contexts, there is a natural tendency to pursue a one-size-fits-all solution to privacy. Unfortunately, this overlooks a crucial insight from decades of privacy research: privacy is inherently contextual. In this paper, we argue that existing work in prompt privacy is drifting towards attempts to design a unified framework, which is highly misleading and can foster a false sense of security. Furthermore, we argue that even future work in this direction needs to wrestle with fundamental tensions across various desirable properties of prompt privacy. Thus, such continued attempts towards a unified framework need to be reconsidered. We recommend three critical paths forward: application-specific deployment of \llm, context-specific evaluation of privacy frameworks, and better privacy literacy.
\end{abstract}

\section{Introduction}
\label{sec:introduction}

\glsreset{llm}

\llm have become embedded in daily workflows, ranging from coding assistants to legal advisors and medical support tools~\cite{zhao2025surveylargelanguagemodels, minaee2025largelanguagemodelssurvey, surveyeval}. With their growing capabilities, \llm have fundamentally changed how users interact with third-party systems \cite{mireshghallah2024trustbotdiscoveringpersonal, touvron2023llamaopenefficientfoundation}. This shift has led to a significant increase in the amount of sensitive information users share, whether due to the personalized and open-ended nature of conversations, better interactivity, or simply a lack of privacy literacy~\cite{274574, zhang2024s, mireshghallah2024trustbotdiscoveringpersonal}.

Unsurprisingly, concerns around privacy in LLMs are on the rise \cite{du2025privacyfinetuninglargelanguage}.
Several solutions have been proposed, including anonymization-like techniques, differential privacy, cryptography, and various attempts of operationalizing the principles of contextual integrity~\cite{zhang2025adanonymizerinteractivelynavigatingbalancing, hong2024dpoptmakelargelanguage, shao2025privacylensevaluatingprivacynorm, dp-forward, hideandseek,mireshghallah2024llmssecrettestingprivacy,10.5555/1822585}.
The goal, it seems, is a user-facing, plug-and-play framework that can be seamlessly integrated into any chatbot.


This evolution in prompt privacy literature reflects a broader trend toward unified solutions. Promises like \textit{``Secure your AI. Everywhere it matters"} \cite{prompt_security}, \textit{``Privacy firewall for ChatGPT prompts"} \cite{privacygpt_chrome_ext}, \textit{``comprehensive tool designed to fortify the security of LLMs''} \cite{protect2024llm}, 
etc., position these solutions as universal safeguards against all privacy risks. 
Clearly, as \llm become general-purpose, research in prompt privacy attempts to follow suit.
However, a one-size-fits-all approach to prompt privacy is deeply flawed.

\begin{figure}[t]
    \centering
    \includegraphics[width=0.8\linewidth]{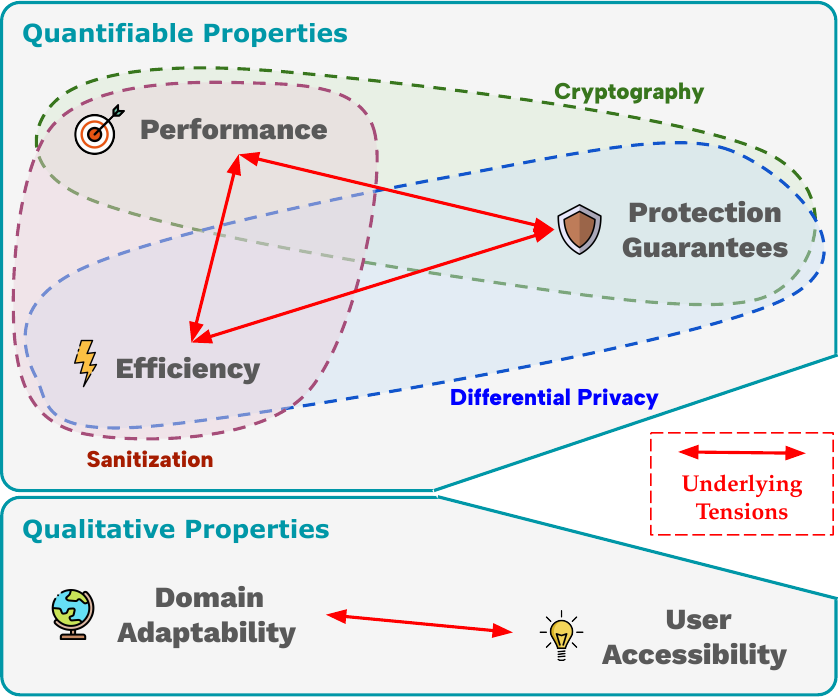}
    \caption{Five desirable properties of a prompt privacy framework (\cref{sec:background}), coverage of various existing frameworks in the literature (\cref{sec:all_frameworks}), and several non-trivial underlying tensions (\cref{sec:fundamental}). 
    }
    \label{fig:tensions}
\end{figure}


Building on existing privacy literature~\cite{delgado2025tapsi,dibia2024sok,iravantchi2025sok,murtagh2018usable,park2023nobody}, we identify five key properties for prompt privacy (see Figure \ref{fig:tensions}): three quantifiable (protection guarantees~\cite{tong2025vulnerabilitytextsanitization,xin2025falsesenseprivacyevaluating}, performance~\cite{sun2024deprompt,hideandseek}, and efficiency~\cite{privacypressurvey,gim2024confidential}) and two qualitative (domain adaptability~\cite{mireshghallah2024llmssecrettestingprivacy,hannah22} and user accessibility~\cite{prompt_security,privacygpt_chrome_ext,zhang2025adanonymizerinteractivelynavigatingbalancing}). 
We uncover several fundamental tensions, both between quantifiable properties (balancing protection, performance, and efficiency) and between qualitative properties (inherent accessibility concerns for adaptable frameworks capable of handling various privacy requirements). Hence, a unified solution to prompt privacy is not a purely technical problem that can be solved in the future.


\begin{figure*}[t]
    \centering
    \includegraphics[width=0.9\linewidth]{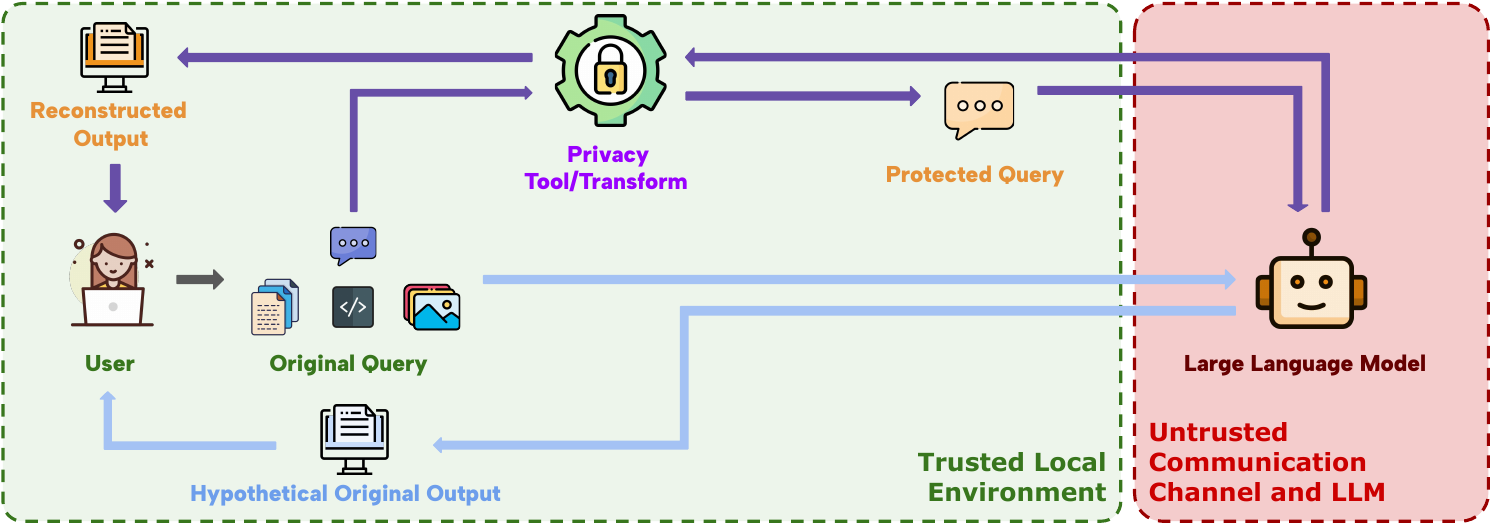}
    \vskip 0.1in
\scriptsize
\begin{tabular}{>{\raggedleft\arraybackslash}p{3.2cm} p{13cm}}
\toprule
\textbf{User} & The individual or entity interacting with the \llm. \\[0.2em]
\textbf{Original Query} & The original input the user intends to send to the \llm. This includes both the language prompt and any auxiliary content such as documents, images, code, etc. \\[0.2em]
\textbf{Privacy Tool/Transform} & The privacy tool responsible for protecting the original query and removing sensitive information, producing a protected query. In some cases, it also post-processes the \llm’ response to reconstruct or recontextualize it before returning it to the user. \\[0.2em]
\textbf{Protected Query} & The sanitized query generated by the privacy tool/transform. \\[0.2em]
\textbf{LLM} & The target \llm to process the query. \\[0.2em]
\textbf{Protected Output} & The response generated by the \llm when given the protected query as input. \\[0.2em]
\textbf{Reconstructed Output} & The final response presented to the user, which may be the protected output or a refined version of it, depending on the privacy tool/transform. \\[0.2em]
\textbf{Hypothetical Original Output} & The response generated by the \llm if given the original query as input. \\[0.2em]
\textbf{Trusted Local Environment} & The environment that the user trusts. This need not be just the user’s device; for example, if a trusted party, say the bank, intermediates access to an \llm, the trusted environment includes both the user’s device and any infrastructure under the control of the bank. \\[0.2em]
\textbf{Untrusted Communication Channel and \llm} & The communication channel through which the query and the output are transmitted and the \llm itself, both assumed not to be trusted and can contain malicious adversaries. \\
\bottomrule
\end{tabular}

    \caption{Prompt Privacy Pipeline and a Glossary of terms used throughout the paper. The user interacts with a privacy tool/transform before their query is sent to the \llm. All privacy interventions occur within a local trusted environment, ensuring that only the protected query is transmitted. We assume both the \llm and the communication channel can have malicious actors. The final reconstructed output is compared against the hypothetical output of the original query to make sure the LLM utility is preserved.}
    \label{fig:scope}
\end{figure*}

\textit{\textbf{We argue that privacy in context is being overlooked in existing research on prompt privacy (with spillover effects in the industry), drifting instead towards a unified framework. However, no single framework for prompts can simultaneously achieve all desirable properties of privacy, due to non-trivial underlying tensions. Instead, presenting such a unified framework will mislead and jeopardize end users.}}




We structure our argument as follows.
We begin by systematically reviewing the prompt privacy literature, highlighting the drift towards a unified framework and the tendency to overlook privacy in context (\cref{sec:unified}). We then identify five desirable properties for prompt privacy frameworks: protection guarantees, performance, efficiency, domain adaptability, and user accessibility (\cref{sec:background_desirable}), and use them to closely examine four common approaches to prompt privacy: sanitization, differential privacy based techniques, cryptographic methods, and attempts to operationalize contextual integrity, highlighting their limitations (\cref{sec:all_frameworks}).
Building on these limitations, we finally identify several underlying tensions between these desired properties, which pose significant non-trivial challenges even for future developments (\cref{sec:fundamental}).
We conclude with three key recommendations: {\bf (1)} prioritizing application-specific deployment of \llm, {\bf (2)} context-specific evaluations of prompt privacy tools, and {\bf (3)} better privacy literacy and awareness for the end users (\cref{sec:solution}).

\section{Anatomy of a Prompt Privacy Framework}
\label{sec:background}


To establish consistent terminology and clearly mark the scope of our work, we present the prompt privacy pipeline and its various components in Figure~\ref{fig:scope}.

By prompt privacy, we refer exclusively to the privacy concerns associated with a user's query to third-party \llm. This includes not only the prompt by the user, but also any additional auxiliary information sent as part of the query, such as documents, images, code, demonstrations, etc. The adversary might be present in either the communication channel or the third-party \llm themselves, as long as the target for information leakage is the user query.

For the same reason, any form of information leakage where the privacy target is not the query itself, even if it happens during inference, is outside the scope of our discussion.
This includes extensive literature on extracting training data from \llm during inference~\citep{274574,more2024realisticextractionattacksadversarial}, model inversion attacks through adversarial prompts~\citep{qu2025prompt,zhao2025survey}, or jailbreaking and backdoor behaviours~\citep{yi2024jailbreak,zhou2025survey}, among others.
\section{Drift Towards a Unified Framework}
\label{sec:unified}

\begin{figure*}[t]
    \centering
    \includegraphics[width=0.55\linewidth]{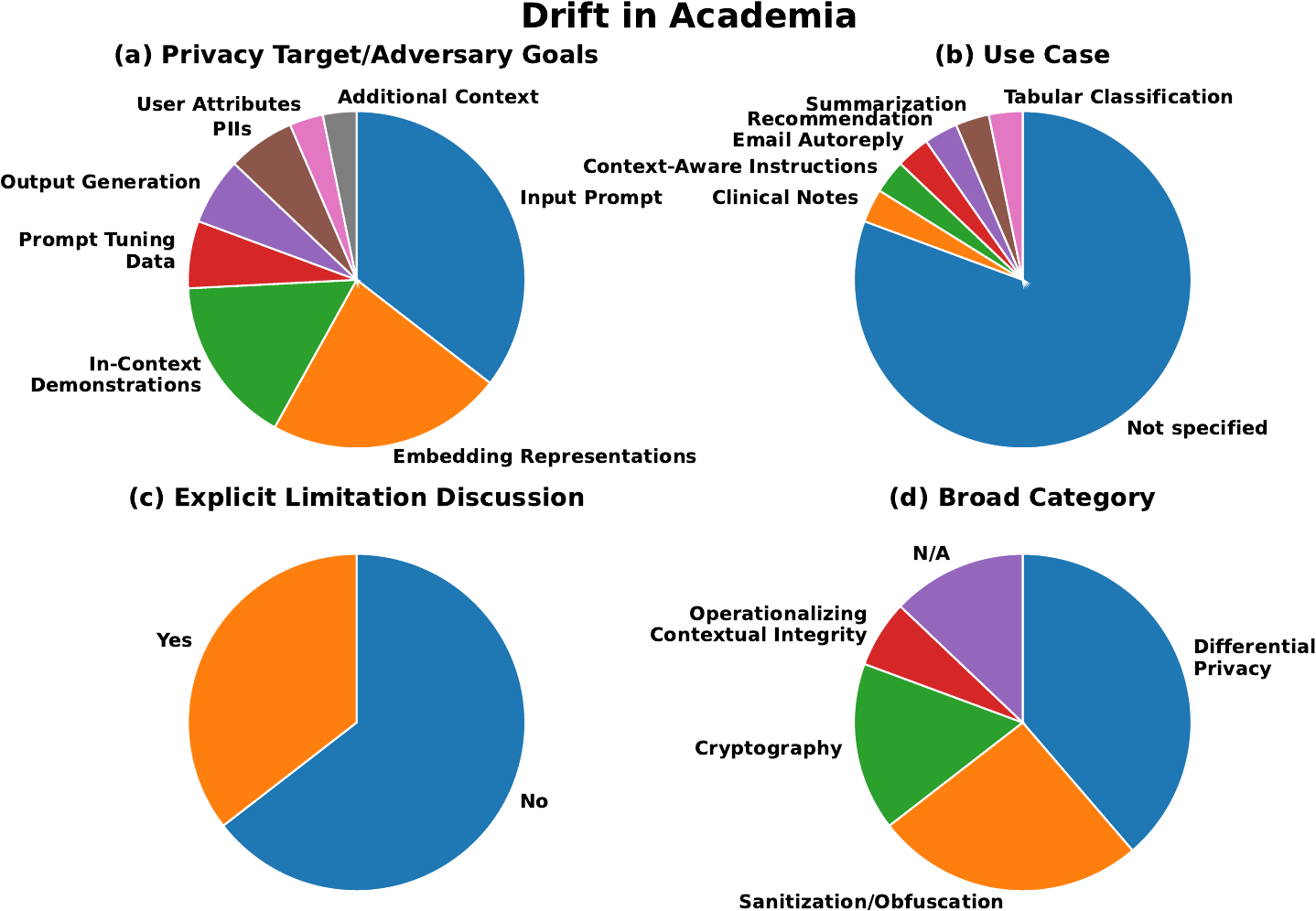}
    \hfill
    \includegraphics[width=0.33\linewidth]{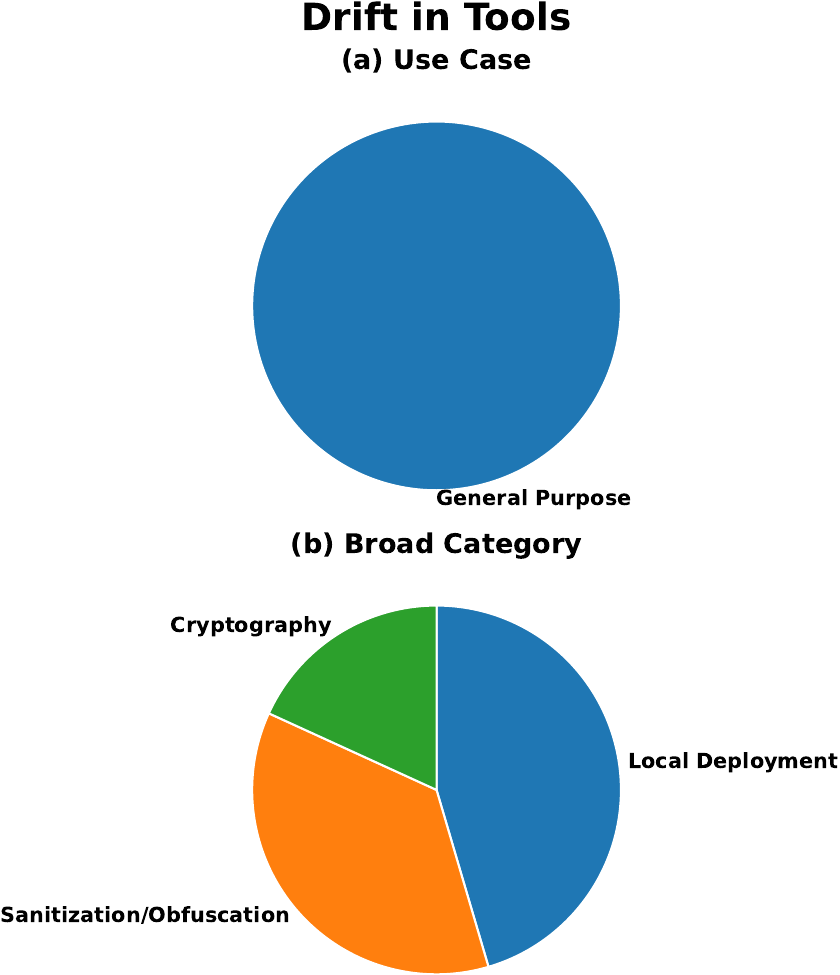}
    \caption{\textbf{(Left)} Detailed annotations of literature in prompt privacy, highlighting the evidence of a drift away from privacy in context. \textbf{(Right)} Detailed annotations of publicly available tools for prompt privacy, highlighting similar trends of drift as the literature.}
    \label{fig:drift}
\end{figure*}

Our first argument builds on a drift toward unified, one-size-fits-all privacy frameworks for \llm in both academia and actual tools. While concerns about the limitations of privacy without context have appeared in prior literature \cite{aunon2024evaluation,delgado2025tapsi,dibia2024sok,iravantchi2025sok,murtagh2018usable,park2023nobody}, the phenomenon of unified frameworks for prompt privacy during inference with \llm is distinct due to the general-purpose nature of the underlying system. In this section, we describe this drift and how it differs from prior treatments of privacy, supported by a systematic study.

\subsection{Prompt Privacy as a Distinct Phenomenon}

The argument that ``privacy is contextual'' is well established in the literature~\citep{nissenbaum2004privacy}. In this paper, however, we focus specifically on prompt privacy, i.e., the privacy of natural language inputs to \llm and an average user’s main privacy concern when interacting with chatbots. We argue that prompt privacy differs fundamentally from previous discussions of privacy in the classical ML literature. Before the advent of general-purpose \llm, systems were designed for specific applications, each requiring its own explicit and contextually appropriate privacy framework. While unified privacy frameworks were still critiqued in these settings \cite{aunon2024evaluation,delgado2025tapsi,park2023nobody}, each application still had to make its own choice of framework, and the criticism was against this choice being made lazily by the developer. In other words, when deploying a model, the context is fixed and it has been shown that the choice of privacy guardrails should depend on it.

But \llm break this paradigm. \llm today are simultaneously responsible for a vast and ever-growing range of tasks, for instance, from simple grammatical assistance to sensitive medical or legal guidance. In this setting, the challenge is no longer the lazy choice to privacy. Instead, the workflow of \llm inherently encourages a single privacy framework that can be applied across a vast and heterogeneous space of user interactions, making it an extremely attractive and often the most accessible option. 
This creates a powerful drift toward attempts to create unified privacy solutions, overlooking important lessons from privacy literature.


\subsection{Evidence of the Drift in Academic Literature}



To better understand this drift, we conduct a systematic study of the existing literature on prompt privacy.
We start with four recent surveys covering a significant breadth of research on privacy in \llm: \citet{edemacu2025privacy,li2023privacy,chen2025survey,smith2023identifying}. First, a broader selection of papers was made from these surveys, identifying categories that contain papers on prompt privacy, resulting in a total of 228 papers after removing duplicates. All papers were then checked manually to see their fit within the scope of our work. After filtering papers that do not focus on prompt privacy, we were left with a total of 31 papers.
Each paper was then read thoroughly and annotated to identify the privacy target, the specific use case (if any), the broader category of privacy-preserving technique employed, and whether the paper included an explicit discussion of its limitations in protecting privacy.
More details about paper selection, filtering, and annotation guidelines, as well as the annotations across all 31 papers, are present in \cref{sec:app_unified}.

Figure \ref{fig:drift} (Left) summarizes key trends across the annotated papers. We find more than half of the papers target privacy for either the entire input prompt or its embedding representations (Figure \ref{fig:drift} (Left)(a)), while over three-fourths propose frameworks intended to generalize across all \llm use cases rather than addressing specific applications (Figure \ref{fig:drift} (Left)(b)). Moreover, more than half of the papers provide no discussion of potential limitations or failure cases (Figure \ref{fig:drift} (Left)(c)). Collectively, these findings reveal a clear drift in the literature toward unified prompt privacy frameworks, often without sufficient contextual grounding. Irrespective of whether this is driven by the pressure to publish and thus not being truthful about limitations~\citep{nosek2012scientific} or simply a lack of attention given to privacy in context, the drift towards unified frameworks is clearly present in existing literature and can even have spillover effects as discussed next.

\subsection{Evidence of the Drift in Prompt Privacy Tools}

In addition to academic literature, we also study various publicly available tools that are (presumably) deployed by real users to protect privacy when interacting with \llm. We started our list from \href{https://github.com/pluja/awesome-privacy}{github.com/pluja/awesome-privacy}, and then expanded the search on the web using the terms `prompt privacy' and `LLM inference privacy'. We found a total of 11 different prompt privacy tools from various sources. Each tool was annotated to identify the use case and the broader category of privacy-preserving technique. More details about the search, as well as the annotations across all 11 tools, are present in \cref{sec:app_unified}.

The trends are collected in Figure \ref{fig:drift} (Right). Interestingly, all 11 tools described their use case as protection during general purpose chatbot usage, and we found no tools designed for specialized contexts (Figure \ref{fig:drift} (Right)(a)). Close to half the tools prioritized local deployment of \llm as the solution to privacy concerns, and the rest relied on either sanitization techniques or deployment on \tee using cryptography (Figure \ref{fig:drift} (Right)(b)). This surfaces another trend worth noting: while a significant portion of literature focuses on differential privacy, these techniques are not present in real-world tools, highlighting the power of accessibility in deciding what gets adopted.
\section{Desirable Properties of Prompt Privacy}
\label{sec:background_desirable}

Now that we've shown how privacy in context is being overlooked, we move to our second argument: a unified framework can never achieve all desirable properties of a prompt privacy framework, and instead will mislead and harm end users. We built this argument in three steps: first, in this section, we identify key desirable properties of prompt privacy. Next, we study various existing prompt privacy frameworks and identify their limitations in achieving these properties. Finally, we build on the discussion of various limitations to identify deeper underlying tensions in prompt privacy.

\subsection{Quantifiable Properties}
We define three quantifiable properties fundamental to any privacy framework, and not just prompt privacy.


\textbf{Protection Guarantees.} 
Central to a privacy framework is its ability to protect against information leakage, i.e., limiting the information that an adversary can extract through the protected query. 
Depending on the context, the threat model, and the information available to the attacker, different strengths and nature of privacy protection might be desirable \cite{tong2025vulnerabilitytextsanitization, papadopoulou2022neural, xin2025falsesenseprivacyevaluating,hong2024dpoptmakelargelanguage, li2025dpgtrdifferentiallyprivateprompt, lu2023bumblebee}.
For instance, protection against honest-but-curious service providers requires different mechanisms than protection against malicious attackers.
Clearly, a framework’s ability to provide protection guarantees against the relevant adversary is a critical property.

\textbf{Performance.} 
While protection guarantees are important, a good privacy framework must also match the performance of accessing the \llm directly, otherwise the switch to a privacy framework would incur utility costs. Several different ways to quantify performance have been proposed, such as the semantic equivalence of the reconstructed and original output, the actionability of the reconstructed output, or the preservation of factual accuracy and consistency, among others~\cite{sun2024deprompt,hideandseek,hong2024dpoptmakelargelanguage}. Broadly, a good privacy framework should maintain output fidelity.

\textbf{Efficiency.} 
Efficiency relates to the computational cost and latency of the privacy tool. A good privacy framework should minimize additional cost, avoid significant latency, and scale effectively~\cite{privacypressurvey,gim2024confidential}. By keeping the processing lightweight and responsive, one can ensure that privacy protections do not hinder the use of \llm. Together, performance and efficiency focus on preserving usability, thus maintaining the practical viability of the privacy framework.

\subsection{Qualitative Properties}
Not all desirable properties of a privacy tool are easily quantifiable. We describe two important qualitative properties.

\textbf{Domain Adaptability.}
Real-world scenarios demand privacy frameworks that can flexibly adapt to diverse domains, even more so with the ever-growing list of use cases for \llm~\cite{mireshghallah2024llmssecrettestingprivacy}. This can take many forms, such as dynamically adjusting to different expectations of privacy protection and types of adversaries, or being able to handle specialized language or conversational formats~\cite{hannah22, privacypressurvey}.
Thus, the ability to adapt across various domains and LLM use cases is essential for an effective unified privacy framework.

\textbf{User Accessibility.} 
The successful adoption of any privacy framework depends on its accessibility to end users. This includes both the transparency of the framework as well as the cognitive burden of incorporating it into everyday workflows. When users can understand how their queries are transformed, they can verify that meaning has been preserved or debug cases where outputs are suboptimal, which can empower them to maintain appropriate protection.
Notably, most widely adopted privacy tools tend to employ \textit{sanitization} \cite{prompt_security,privacygpt_chrome_ext,chong2024casperpromptsanitizationprotecting, zhang2025adanonymizerinteractivelynavigatingbalancing} (Figure \ref{fig:drift} (Right)(b)), a privacy framework that prioritizes accessibility. The appeal of a low barrier of adoption for the end user cannot be overstated.

\textit{Note that the qualitative properties operate at a higher level than the quantifiable properties. For example, domain adaptability may be understood as a framework’s ability to provide appropriate protection guarantees while maintaining performance and efficiency, across diverse domains. Similarly, user accessibility is the extent to which a framework makes the relationship between performance, efficiency, and privacy understandable and controllable for the end user.}

\section{User Profiles as Motivating Examples}
\label{sec:examples}

To illustrate our arguments, we introduce three motivating examples, which we will reference throughout the paper.

\textbf{Avery’s Marathon Preparation.}
Avery wants to leverage an LLM to design a personalized fitness journey and prepare for a marathon. They aim to develop dietary plans, identify exercises, and create a comprehensive strategy to run the marathon in six months. As a meticulous researcher, Avery is not bothered about the accuracy of the LLM’s suggestions, since they will verify the details independently. Their primary goal is to generate a quick initial draft of the plan. However, Avery is concerned about targeted ads if their chat history gets leaked or is sold by the third-party LLM.

\textbf{Sasha's Work Companion.}
Sasha's company has recently rolled out a work companion using a third-party LLM API to assist employees with routine tasks such as summarizing reports, creating presentations, and more. Sasha plans to use this tool when working with confidential documents, so strong privacy protections are critical. A tool that requires constant supervision would be counterproductive, so it needs to be highly accurate. Sasha's company is managing the costs, and the tool will run on a centralized infrastructure, while using the third-party LLM API underneath. Note that the trusted local environment here includes the company infrastructure, but not the third-party LLM.

\textbf{Tao's Writing Assistant.}
Tao is a non-native English speaker who wants to use an LLM-based writing assistant to help with personal emails and other forms of communication. Since these interactions are mostly casual, Tao is not overly concerned about privacy risks. However, they would like to ensure that no unnecessary identifying information is shared with the LLM. For Tao, accuracy and speed, allowing seamless integration of the tool into their daily life, are more important than strict privacy guarantees.
\section{Drawbacks of Existing Frameworks}
\label{sec:all_frameworks}


To understand why a single framework cannot simultaneously achieve all desirable properties of prompt privacy, we first enumerate the existing limitations of the four prompt privacy solution categories present in the literature (Figure \ref{fig:drift} (Left)(d)): sanitization, differential privacy based techniques, cryptographic methods, and various attempts at operationalizing the theory of contextual integrity. 


\textit{Note: While sanitization and cryptographic methods are concrete techniques, differential privacy is a mathematical concept that can be implemented in many different ways, and contextual integrity is a social theory of privacy. Neither of them directly refer to prompt privacy techniques. However, we overload these terms in the paper to facilitate readability, instead of repeatedly clarifying these differences. We use the term differential privacy to refer to a broad class of techniques that share the common property of providing formal differential privacy guarantees. Similarly, our discussions of contextual integrity focus on techniques that attempt to operationalize the social theory of contextual integrity.}

\begin{figure*}[t]
    \centering
    \includegraphics[width=0.75\linewidth]{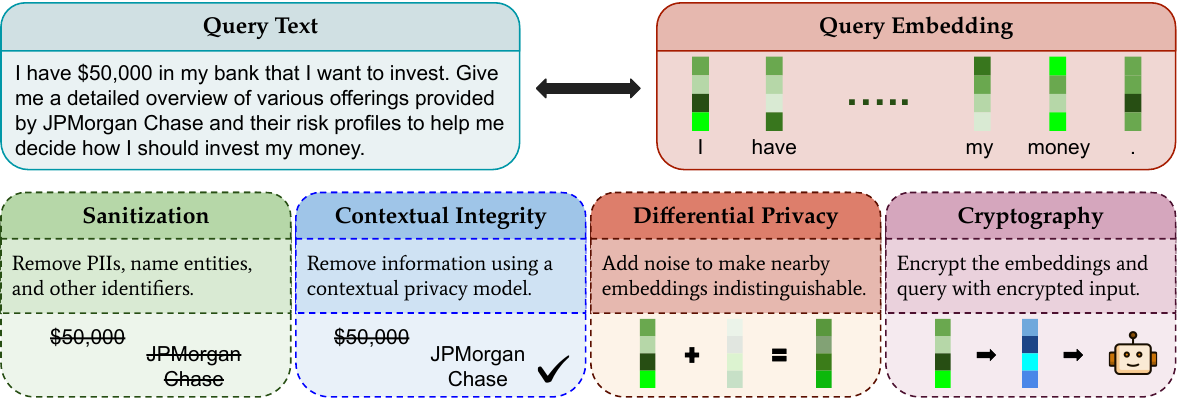}
    \caption{Four commonly adopted frameworks for prompt privacy: sanitization, differential privacy, cryptography, and contextual integrity.}
    \label{fig:frameworks}
\end{figure*}

\subsection{Sanitization: A Misplaced Sense of Safety}
\label{sec:sanitization_sub}

Sanitization refers to techniques that remove, mask, or generalize parts of the prompt in an effort to reduce exposure of sensitive information, as illustrated in Figure \ref{fig:frameworks}. These methods work at the surface level, making them efficient without degrading performance, but they lack rigorous protection guarantees. Sanitization often relies on lexical patterns and is thus highly accessible for end users.

\textbf{Lexical Sanitization misses Semantic Context.}
Lexical sanitization \cite{hedegaard2009lair,papadopoulou2022neural, albanese2023text, sanchez2016csanitized} is a technique that focuses on sanitizing specific words or phrases from a prompt, without considering the semantic meaning or the broader context. This encompasses common approaches such as \pii removal and \ner-based methods. However, these techniques are frequently criticized for overlooking the complex linguistic and relational cues present in natural language \cite{guo2024benchmarkinglinguisticdiversitylarge,sinha2022languagemodelacceptabilityjudgements}. Moreover, since lexical sanitization methods rely on predefined lists of words or categories, they often fail to generalize across domains where sensitive information can take many different forms. As a result, what qualifies as sensitive in one setting may go undetected in another, limiting their adaptability across domains \cite{mireshghallah2024llmssecrettestingprivacy}.


\textbf{Lack of Protection Guarantees.}
Recent works have revealed that the effectiveness of sanitization in removing sensitive information is often overestimated. Black-box attacks on \llm show adversaries can frequently reconstruct and infer sensitive information, even after applying sanitization or de-identification, by exploiting residual semantic cues and contextual patterns \cite{carpentier2025preemptingtextsanitizationutility,hong2024dpoptmakelargelanguage, li2025privacypreservingprompttuninglarge,tong2025vulnerabilitytextsanitization, xin2025falsesenseprivacyevaluating}.
Such findings highlight that surface-level masking is rarely sufficient for robust privacy protection, especially in real-world scenarios with motivated adversaries \cite{li2025dpgtrdifferentiallyprivateprompt}. Hence, while sanitization can create a sense of privacy by removing certain words from the sentence, they have an even greater risk of misleading users \cite{privacypressurvey, tong2025vulnerabilitytextsanitization, papadopoulou2022neural}.


\textbf{The Illusion of Personalization and Control.}
Sanitization tools often advertise user control over how much sensitive information is removed, balanced against task utility. This is a common design choice in user-facing frameworks such as browser extensions \cite{chong2024casperpromptsanitizationprotecting, zhang2025adanonymizerinteractivelynavigatingbalancing} and client-side tools \cite{Zhou2024RescriberSU}.
However, such tradeoff levers are built on the assumption that privacy and performance can be reduced to a single continuum, and that balancing one against the other is both meaningful and tractable \cite{zhang2025adanonymizerinteractivelynavigatingbalancing, Zhou2024RescriberSU}. In reality, privacy risks are highly contextual, shaped by semantic and pragmatic cues~\cite{Zhou2024RescriberSU, zhang2025adanonymizerinteractivelynavigatingbalancing, privacypressurvey}, and even the interpretation of the privacy scale can vary heavily across different users \cite{xin2025falsesenseprivacyevaluating, xiao2023privacymind}. 

\begin{tcolorbox}[colback=green!5!white, colframe=green!75!black, boxrule=1pt, arc=3mm, left=2mm, right=2mm, top=1mm, bottom=1mm]
\textit{Sanitization frameworks offer a surface-level protection from privacy risks, along with illusory control over the balance against utility. While useful in some situations, sanitization cannot withstand motivated adversaries and fails to capture the dynamic, context-sensitive nature of real-world privacy across domains.}
\vspace{0.5em}

Sanitization risks exposing \textbf{Avery} to targeted ads and leaking company data for \textbf{Sasha}, making it unsuitable for them both. It may be an appropriate tool for \textbf{Tao}, providing good performance and efficiency, along with surface-level privacy sufficient for their needs.
\end{tcolorbox}

\subsection{Differential Privacy: Promises and Pitfalls}
\label{sec:dp_sub}

\dpr is a mathematical framework of privacy in data-driven systems, and \dpr guarantees have long been regarded as the gold standard in traditional ML settings~\cite{dwork_algorithmic}. At its core, \dpr guarantees provide a powerful statement: the output of a system is indistinguishably affected by the presence or absence of a single individual data point, achieved by introducing noise. Thus, an adversary with access to the output cannot confidently predict the membership of individuals in the input data.

In prompt privacy, \dpr is applied to the input prompt by injecting random noise through perturbations, ensuring that the perturbed prompt could have plausibly come from many different input prompts~\cite{wucape}, as shown in Figure \ref{fig:frameworks}. Thus, an adversary with access to the perturbed prompt cannot confidently state which input was used to create it.

\textbf{Relative Bounds, Not Absolute Walls.}
\dpr provides relative bounds on information leakage rather than preventing the absolute flow of information. In essence, \dpr ensures that an adversary cannot reliably distinguish between neighboring input prompts. However, if certain information is common across all neighbouring prompts, the adversary can still infer it. For instance, suppose a user asks an \llm about a particular medication. \dpr may perturb the prompt so that the medication name is replaced with that of a similar but different drug. While this prevents the adversary from confidently recovering the original medication, even knowledge of the perturbed alternative still reveals significant information about the user’s potential medical condition.

This tension is not surprising, and in fact, is very well known in the privacy literature as \textit{Dalenius’ Desideratum}, i.e., in simple terms, if you can learn nothing from the data, you've achieved perfect privacy~\citep{dwork_algorithmic}. This trade-off between privacy and utility, thus, is not a limitation of any one particular technique, but it does appear more clearly in \dpr than any other class of privacy-preserving methods.


\textbf{The Structural Limits of \dpr for Prompts.}
Guaranteeing protection under \dpr relies on having a clear definition of a ``secret'' or sensitive record. For language data, and especially for prompts, it is difficult to establish consistent boundaries for what should be protected. Language does not have a universal unit for sensitive information. 

On the one hand, if privacy is defined at the level of an entire conversation, then the resulting guarantees can be overly coarse and impractical, since the mechanism would need to inject prohibitive amounts of noise to provide formal protection, severely degrading utility. However, on the other hand, if privacy is defined at the word level, removing these units rarely hides private information, and does not prevent an adversary from inferring sensitive context from the rest of the prompt or related patterns \cite{hannah22, mireshghallah2024trustbotdiscoveringpersonal}.

\textbf{The Difficulty of Defining Similarity of Prompts.}
Another challenge of defining neighbouring prompts in \dpr lies in choosing a meaningful similarity metric. As similarity in the semantic space is not directly quantifiable, it is often mapped into a different space, such as the embeddings. However, this introduces a gap between the measured similarity and the actual semantic similarity. Consequently, even prompts that appear safe may contain identifying details, undermining the intended privacy guarantees \cite{dp-forward, mattern2022limitswordleveldifferential}.

\begin{tcolorbox}[colback=green!5!white, colframe=green!75!black, boxrule=1pt, arc=3mm, left=2mm, right=2mm, top=1mm, bottom=1mm]
\textit{Differential privacy offers strong guarantees against information leakage. However, its limitations are not just technical obstacles, but reflect a deeper misalignment between the rigidity of differential privacy and the dynamic nature of language.}

\vspace{0.5em}

Performance drop under \dpr would be unacceptable for \textbf{Sasha} and \textbf{Tao}. While \textbf{Avery} could benefit from the protection provided by \dpr, they need literacy of the privacy-utility trade-off to take its advantage.
\end{tcolorbox}

\subsection{Cryptography: A Costly Alternative}
\label{sec:cryptography_sub}

Cryptography in the context of privacy includes a wide range of techniques, including encryption~\cite{acar2018survey}, secret sharing~\cite{du2001secure,zhao2019secure}, zero-knowledge proofs~\cite{fiege1987zero}, oblivious transfer~\cite{rabin2005exchange,yadav2022survey}, etc. Two cryptographic techniques are common in prompt privacy literature: \he and \mpc. \he enables computation directly on encrypted text, as shown in Figure \ref{fig:frameworks}, and decrypts the output in the trusted environment~\cite{zhang2024secpe,deencryptedllm,zimerman2025powersoftmax}. \mpc, on the other hand, distributes the computation across multiple independent parties, ensuring that no single party has access to the complete input~\cite{rathee2024mpc,xu2025breaking}.

\textbf{The Barrier of Non-Linearity.}
\he holds great appeal because it promises the best of both worlds: privacy-preserving encryption and full utility of the model~\cite{acar2018survey}. In practice, however, it is highly constrained. \he is relatively efficient for linear operations, such as the large matrix multiplications. The challenge arises with nonlinear operations like ReLU, softmax, and the attention mechanism itself, which can either make \he extremely slow or require polynomial approximations that quickly become unstable and expensive~\cite{mittal2024improving,xu2025breaking,zhang2023enhancing}. Thus, \he does not scale well for \llm.

\textbf{Specialized Infrastructure Requirements.}
A major barrier to the use of cryptographic techniques for prompt privacy in \llm is the need for specialized infrastructure. \mpc requires multiple non-colluding servers to jointly host and run the model, with heavy inter-server communication during inference. This is at direct odds with the way \llm are deployed today, as centralized services optimized for single-node or tightly clustered GPU execution. Similarly, \llm need to be reimplemented to be \he-compatible, fundamentally changing the serving pipeline itself. Thus, the existing \llm infrastructure does not support these techniques out of the box, and would require substantial re-engineering.

\textbf{Hybrid Solutions and Communication Overhead.}
Recent works have attempted to combine \he and \mpc, taking advantage of the beneficial aspects of each to create hybrid techniques~\cite{xu2025breaking,lu2023bumblebee,pang2024bolt}. While this has provided significant improvements, even under optimistic assumptions of specialized infrastructure and highly tuned cryptographic libraries, hybrid techniques still introduce significant communication costs that accumulate quickly. Consequently, although powerful, they remain far too expensive for \llm.

\begin{tcolorbox}[colback=green!5!white, colframe=green!75!black, boxrule=1pt, arc=3mm, left=2mm, right=2mm, top=1mm, bottom=1mm]
\textit{Cryptographic techniques for prompt privacy promise protection guarantees while maintaining model performance, however, fall short of true practical benefits due to their extremely high computational costs, limiting their use to toy models or shallow networks.}

\vspace{0.5em}

The high cost of cryptographic techniques makes it unacceptable for \textbf{Avery} and \textbf{Tao}. As \textbf{Sasha's} work companion is deployed through the company infrastructure, they could potentially benefit from the guarantees it provides against absolute information leakage.
\end{tcolorbox}

\subsection{Contextual Integrity: A Distant Ideal}
\label{sec:ci_sub}

\ci~\cite{nissenbaum2004privacy,mireshghallah2024llmssecrettestingprivacy} offers a principled way to analyze and adapt privacy norms as technology and society evolve. \ci is the only framework in our discussion that defines privacy not as a static property but as the appropriateness of information flow within a given context, by examining who is sharing information, what is being shared, with whom, for what purpose, and under what transmission principles~\cite{mireshghallah2024llmssecrettestingprivacy,shvartzshnaider2025positioncontextualintegrityinadequately,nissenbaum2004privacy}, an example shown in Figure \ref{fig:frameworks}.

While \ci moves beyond binary public/private distinctions, recent critiques note that research in \llm often ``washes'' \ci in an attempt to operationalize it, adopting \ci terminology while neglecting its core principles \cite{shvartzshnaider2025positioncontextualintegrityinadequately, hannah22}. Thus, translating \ci into practice remains a significant challenge. Several arguments in this section build on \citet{shvartzshnaider2025positioncontextualintegrityinadequately}'s position that \ci is inadequately applied to \llm in the existing literature.

\textbf{Not Just a Special Case of Sanitization.}
Existing attempts to operationalize \ci tend to encode fixed privacy norms, drawing from legal codes, crowdsourced inputs, or author judgment~\cite{shao2025privacylensevaluatingprivacynorm,ghalebikesabi2024operationalizing,cheng2024ci,li2025privaci}. In fact, many approaches that operationalize \ci do so through prompt sanitization~\cite{hartmann2024can,ngong2025protectingusersthemselvessafeguarding,shvartzshnaider2025positioncontextualintegrityinadequately}.
As a result, these tools struggle to adapt to changing social contexts, evolving notions of sensitivity, and the diversity and dynamism inherent in real-world communication \cite{cofone2018privacy, shvartzshnaider2025positioncontextualintegrityinadequately, privacypressurvey}.


\textbf{The Difficulty of Defining Social Norms.}
\ci is frequently reduced to regulatory compliance or minimization, losing the framework's distinctiveness~\cite{shvartzshnaider2025positioncontextualintegrityinadequately}, thus focusing on preventing the leakage of a fixed set of sensitive information or enforcing policy rules. However, true \ci analysis should holistically assess the appropriateness of information flow in context, considering roles, values, and functions, not merely whether ``private'' information was shared \cite{shvartzshnaider2025positioncontextualintegrityinadequately, yang2013memorization, barkhuus2012mismeasurement}.
Laws and regulations, while influential, do not always reflect lived or moral norms in society, and privacy preferences collected from users may fail to capture the collective or ethical nature of privacy that \ci intends to protect \cite{neel2023privacy}.


\begin{tcolorbox}[colback=green!5!white, colframe=green!75!black, boxrule=1pt, arc=3mm, left=2mm, right=2mm, top=1mm, bottom=1mm]
\textit{Contextual integrity aims to study the dynamic nature of information flow in prompt privacy. However, translating this into a practical tool remains a challenge, particularly without losing the very fluidity that makes it valuable. As \citet{shvartzshnaider2025positioncontextualintegrityinadequately} argue, contextual integrity is a social theory, and requires more than just purely algorithmic solutions.}

\vspace{0.5em}

The high barrier to entry makes contextual integrity an unlikely choice for \textbf{Tao}. For others, it could be effective, for example, keeping \textbf{Avery} protected from targeted ads while still delivering a useful strategy, or preventing \textbf{Sasha} from sharing certain confidential documents. However, such context-specific tools should already exist and be accessible to the end users.
\end{tcolorbox}
\section{Underlying Tensions in Prompt Privacy}
\label{sec:fundamental}



We have examined the existing shortcomings and challenges across various techniques for prompt privacy.
Building on this, we now complete our second argument by identifying several underlying non-trivial tensions inherent to prompt privacy for \llm that would prevent the development of a unified framework even in the future.

\subsection{The Privacy-Utility-Efficiency Trade-off}

We start by briefly acknowledging the fundamental privacy–utility–efficiency trade-off even beyond the context of \llm. For data to remain useful, it must retain some information, which inevitably leads to some degree of privacy leakage. At the same time, identifying the optimal balance between privacy and utility is itself highly complex, and achieving increasingly refined trade-offs often comes at a cost, thereby reducing efficiency. These tensions have dominated the privacy literature for decades~\citep{dwork_algorithmic,makhdoumi2014information}, so we do not elaborate on them further here. Instead, we focus on tensions that arise specifically in prompt privacy.

\subsection{Natural Language or Embedding Representations}
Regardless of the underlying model, \nlp systems transform language into vector representations through embeddings. This transformation from natural language to representation creates an inherent disconnect that generates tension between two critical dimensions: protection guarantees and performance.

The core challenge stems from differences in defining ``similarity'' across the two spaces. Semantic similarity in language space may not always align with similarity in the representational space (\cref{sec:dp_sub}), affecting both protection guarantees and utility. Privacy frameworks that focus on the representations, such as \dpr (\cref{sec:dp_sub}) and cryptography (\cref{sec:cryptography_sub}), can provide formal guarantees, but only in the representation space. To truly prevent sensitive information flow, these techniques need to `over-protect' in the representational space, to ensure appropriate protection in the language space, at the cost of degraded performance (\dpr, \cref{sec:dp_sub}) or reduced efficiency (cryptography, \cref{sec:cryptography_sub}).
On the other hand, frameworks that operate in the language space, such as sanitization (\cref{sec:sanitization_sub}), can maintain performance but lack any protection guarantees.

This is more than a trivial limitation of existing solutions. There is a tension between stronger guarantees in the representational space (at the cost of performance or efficiency) or better performance and efficiency in the language space (at the cost of formal guarantees). Privacy preserving frameworks must choose between these two distinct spaces that are likely to remain fundamentally irreconcilable, even in the long term (the \textit{`symbol grounding problem'} or in its modern form, the \textit{`vector grounding problem'}~\citep{harnad1990symbol,mollo2023vector}).

\subsection{User Accessibility or Domain Adaptability}

The tension between user accessibility and domain adaptability emerges from the remarkable breadth of applications for which \llm are now employed. This diversity of use cases creates vastly different privacy requirements across domains (\cref{sec:sanitization_sub}, \ref{sec:ci_sub}). 
Unfortunately, a framework capable of truly adapting to this variety of domains and covering all different requirements necessarily demands significant privacy literacy from users. Thus, the accessibility and ease of use of a privacy framework stand in fundamental opposition to the comprehensive adaptability required for \llm. This creates an unavoidable choice: privacy frameworks can either be accessible and easy to use, or they can be highly adaptable across domains, but not both.

Accessible frameworks like sanitization (\cref{sec:sanitization_sub}) require minimal cognitive overhead. However, such frameworks struggle with domain adaptability due to changing specialized terminology across fields, evolving social norms, and varying definitions of what constitutes sensitive information in different contexts (\cref{sec:sanitization_sub}). On the other extreme, contextual integrity is specifically designed for cross-domain adaptability (\cref{sec:ci_sub}). However, it demands domain-specific expertise, an understanding of how privacy situations are constructed, and how sensitivity is defined in various contexts. 

Again, the tension is not limited to current approaches, but is instead inherent to prompt privacy. This mirrors the popular \textit{`convention vs configuration'} argument, fundamental to the world of product design~\citep{rails_doctrine}.
As \llm continue to expand into new domains with unique privacy requirements, a framework attempting to maintain both accessibility and comprehensive domain adaptability will face difficult trade-offs between these competing demands.

\section{The Complete Argument}
Through systematic analysis, we showed that privacy in context is being overlooked in the age of general-purpose \llm, both in academic research and in the privacy tools adopted by users (\cref{sec:unified}). We then argued that this shift toward unified privacy frameworks is fundamentally problematic. By examining existing prompt privacy techniques, we identified the limitations of each approach and showed that no current method can simultaneously satisfy all desirable properties of a prompt privacy framework (\cref{sec:all_frameworks}). Looking beyond existing solutions, we then identified several deeper, non-trivial tensions that suggest these limitations are unlikely to disappear even in the future (\cref{sec:fundamental}). Thus, treating privacy in \llm as one uniform problem is misleading and risks causing real harm by giving users an inaccurate understanding of the protections their chosen techniques actually provide.

\section{Alternative Views}

\begin{figure*}[t]
    \centering
    \includegraphics[width=0.8\linewidth]{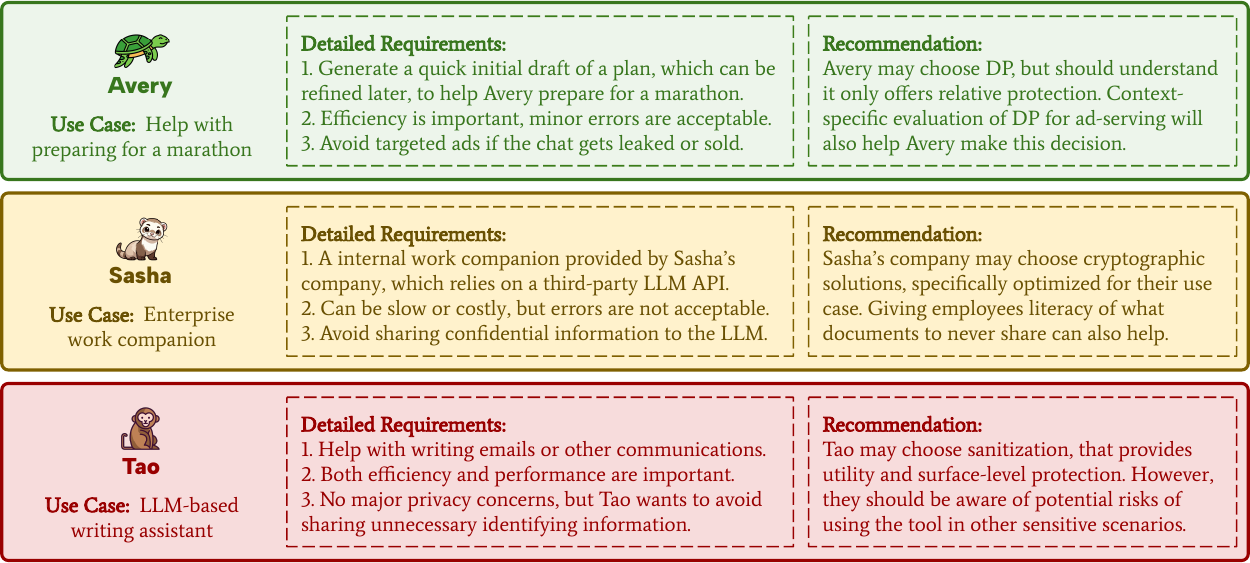}
    \caption{Motivating use cases as defined in \cref{sec:examples}, along with our recommendations based on the preceding discussions in the paper.}
    \label{fig:cases}
\end{figure*}


Below, we briefly discuss three potentially alternative views.

\textbf{Unified Frameworks are Necessary, even if Imperfect.}
A reasonable argument against our paper is that unified frameworks, though imperfect, are necessary given the rapid adoption of \llm. If the field starts to focus exclusively on context-specific solutions, this may create more harm, as users will face an overwhelming number of choices based on the context, and might end up not choosing at all. At least an imperfect framework is better than none.

Indeed, a baseline framework can be important, and we don't argue that research in this direction should be abandoned. However, an imperfect framework is not necessarily preferable to having none. In some cases, such frameworks may create a false sense of security, leading users to become complacent and more willing to share sensitive information than they would if they were aware of potential risks. This naturally leads to our second objection to this argument: even when such baselines or ``unified'' frameworks are implemented, their limitations must be clearly and explicitly communicated, something heavily undervalued in the current prompt privacy literature (Figure \ref{fig:drift} (Left)(c)).

\textbf{Unified Frameworks are Stepping Stones.}
An argument can be made that prompt privacy is a young field and, like any emerging area of research, unified frameworks are merely a stepping stone. The field will naturally diversify as it matures, and from this perspective, privacy in context is not being overlooked but is simply underdeveloped.

We do not aim to make any claims on how the field might evolve. However, we argue that the rapid deployment of \llm in real-world applications has already translated research norms into commercial products. Many tools now market themselves as a ``one-stop'' privacy solution~\citep{prompt_security,privacygpt_chrome_ext,protect2024llm}, reflecting and reinforcing the same sensibilities that we criticize in the academic literature (Figure \ref{fig:drift} (Right)(a)). Thus, we cannot wait for the field of prompt privacy to naturally grow towards privacy in context, as it is already shaping the development of real-world tools, and an intervention is both necessary and urgent.

\textbf{Context can be Learned.}
We do not provide an extensive counterargument here, as \citet{shvartzshnaider2024llm} discusses this issue in depth: contextual frameworks cannot be standalone technical solutions, but need to be sociotechnical approaches that require broader ecosystem-level efforts.
\section{Where do we go from here?}
\label{sec:solution}


To carve a path forward, we bring the reader’s attention back to our motivating examples in \cref{sec:examples}. Throughout our discussion, we found that even though unified frameworks are out of reach, certain frameworks can benefit these users in concrete scenarios. We summarize these solutions in Figure \ref{fig:cases}. 
Two key points deserve emphasis and form the basis of our recommendations: first, effective application- and context-specific solutions are possible even without a universal framework; second, improving end users’ privacy literacy is essential, both to help them select the appropriate solution and to understand its limitations.

\subsection{Context-Specific Evaluations}

The first step toward choosing a privacy tool is recognizing the context-specific requirements of the application. Understanding what the user needs, what trade-offs they are willing to accept, and which privacy risks are most relevant helps guide the selection of the most appropriate tool. Theories such as contextual integrity~\cite{nissenbaum2004privacy} can be particularly useful in this process. 
The \textit{``Detailed Requirements''} in Figure \ref{fig:cases} illustrates this exercise well.

Once the requirements are identified, the next challenge is assessing which privacy framework performs well in that setting. For instance, based solely on the properties of \dpr, it is unclear whether it can provide sufficient protection against targeted ads for Avery. Similarly, the boundary of when sanitization can be problematic in email writing for Tao remains poorly understood. While many works focus on improving privacy frameworks, evaluating these systems in specific contexts is rare.
As we saw in our systematic study, less than a quarter of the work in prompt privacy literature ties itself to a specific application (Figure \ref{fig:drift} (Left)(b)).


\textit{We strongly recommend that the community move toward context-specific evaluation of privacy frameworks, to audit their strengths, limitations, and suitability.}

\subsection{Application-Specific Deployment of \llm}

Another way of enforcing context-specific privacy protections is to separate different uses of \llm. Despite their general-purpose capabilities, separating applications with substantially different privacy requirements can provide meaningful benefits. Consider the recent releases of \textit{ChatGPT Health}~\citep{openai2026chatgpthealth} and \textit{Claude for Healthcare}~\citep{anthropic2026claudehealthcare}. By isolating healthcare-related use from other applications and managing their data separately, providers can apply privacy safeguards tailored to the sensitivity of medical information.

The use of LLMs in healthcare raises numerous ethical concerns, of which privacy is only one. We do not endorse the use of LLMs in healthcare settings; rather, we only use this example to illustrate how application-specific deployment can be implemented in practice, even for highly capable general-purpose models. Such siloed deployments enable more context-aware management of privacy risks.


\textit{We recommend that developers move towards application-specific deployment of \llm, wherever reasonably possible, thus avoiding the trap of a unified privacy framework.}

\subsection{Fostering User Awareness and Privacy Literacy}

In the absence of a universal solution and with the search for context-specific tools, it becomes even more important to help end users understand these trade-offs. For instance, while no single tool is ideal for Sasha’s company (cryptography being the closest, though still quite costly), educating employees about what form of confidential information should never be shared with \llm can help reduce information leakage while still providing them with a chatbot companion. Similarly, local deployment of \llm has gained significant popularity to avoid the concerns of privacy. Users with better privacy literacy will be able to switch between locally deployed \llm or third-party \llm, depending on the severity of the context. 


\textit{Empowering users with stronger literacy and accurate mental models of how their data is handled is essential to fostering safer and trustworthy interactions.}

\section*{Acknowledgment}

Funding support for project activities has been partially provided by Canada CIFAR AI Chair, Google award, MITACS, FRQNT, and NSERC Discovery Grants program. We also express our gratitude to Compute Canada for their support in providing facilities for our evaluations.

\bibliography{references}
\bibliographystyle{icml2026}

\newpage
\appendix
\onecolumn

\printglossaries

\section{Systematic Study of Drift Towards Unified Framework for Prompt Privacy}
\label{sec:app_unified}

\subsection{Academic Literature}

We start our systematic study of academic literature by collecting papers from the following four surveys: \citet{edemacu2025privacy,li2023privacy,chen2025survey,smith2023identifying}. For \citet{chen2025survey}, we collect all papers from Tables 1-4, i.e., all papers covered in the survey. For \citet{edemacu2025privacy}, we collect all papers from Table 1, i.e., also all papers covered in the survey. For \citet{li2023privacy}, we collect all papers under the attack stage Model Inference and Others (thus excluding attack stage Model Training) from Table 2. Finally, for \citet{smith2023identifying}, we collect all papers marked for phase In-Context Learning in Table 1, excluding all other phases (pretraining, continued pretraining, and fine-tuning). After removing duplicates, this created a list of 228 papers in total.

Each paper was then manually checked to see if it fit within the scope of our work. In other words, only papers where the privacy target was the input query sent to an LLM during inference, or some part of it, were kept, while the rest were filtered out. This left us with a final list of 31 papers in total.
Each paper was then annotated, answering the following four questions: (a) What is the privacy target for protection, or what kind of adversary is the method protecting against? (b) Is there a specific use case that is being targeted? (c) Is there an explicit discussion of limitations, or contexts in which this framework will not work? and (d) What is the broad category of technique used in the paper (N/A if the paper doesn't propose a defense strategy, but only studies information leakage)? 

In our labeling, we considered the central framing, methodological commitments, and conceptual narrative of each work, not just isolated experimental details or the presence of certain sections. For instance, if a paper claims broad protection against leakage, i.e., does not commit to an adversary, but still defines specific adversaries only for empirical evaluation, we do not count this as protecting against a specific adversary, since this choice does not inform the central narrative of the work. Similarly, the presence of a limitations section alone is not enough, and we specifically look for discussions of when the proposed privacy framework might fail. The final set of annotations for all 31 papers is collected in Table \ref{tab:drift_academic}.

\subsection{Prompt Privacy Tools}

For prompt privacy tools, we started our search from \href{https://github.com/pluja/awesome-privacy}{github.com/pluja/awesome-privacy}, and then expanded the search on the web using the terms `prompt privacy' and `LLM inference privacy'. More specifically, we searched on \href{https://google.com}{google.com}, \href{https://producthunt.com}{producthunt.com}, and \href{https://chromewebstore.google.com}{chromewebstore.google.com}. Each tool was then annotated, answering the following three questions: (a) Is there a specific use case that is being targeted? (b) Is the tool free or paid? and (c) What is the broad category of technique used by the tool? The final set of annotations for all 11 tools is collected in Table \ref{tab:drift_tools}.


\begin{table*}[t]
\centering
\scriptsize
\begin{tabular}{p{4.8cm} p{2.5cm} p{2.5cm} p{2.5cm} p{2.5cm}}
\toprule
\textbf{Title} & \textbf{Privacy Target/Adversary Goals} & \textbf{Use Case} & \textbf{Explicit Mention of Limitations} & \textbf{Technique Category} \\
\midrule

CoGenesis: A Framework Collaborating Large and Small Language Models for Secure Context-Aware Instruction Following~\citep{zhang2024cogenesis} 
& Additional Context 
& Context-Aware Instructions 
& No 
& Contextual Integrity \\

Can LLMs Keep a Secret? Testing Privacy Implications of Language Models via Contextual Integrity Theory ~\citep{mireshghallah2024llmssecrettestingprivacy}
& Input Prompt 
& No 
& Yes 
& Contextual Integrity \\

CipherGPT: Secure Two-Party GPT Inference~\citep{hou2023ciphergpt} 
& Input Prompt 
& No 
& No 
& Cryptography \\

Iron: Private Inference on Transformers~\citep{hao2022iron} 
& Input Prompt 
& No 
& No 
& Cryptography \\

THE-X: Privacy-Preserving Transformer Inference with Homomorphic Encryption~\citep{chen2022x} 
& Input Prompt 
& No 
& No 
& Cryptography \\

A Verified Confidential Computing as a Service Framework for Privacy Preservation~\citep{chen2023verified} 
& Input Prompt 
& No 
& Yes 
& Cryptography \\

SecFormer: Fast and Accurate Privacy-Preserving Inference for Transformer Models via SMPC~\citep{luo2024secformer} 
& Input Prompt 
& No 
& Yes 
& Cryptography \\

Differentially Private Representation for NLP: Formal Guarantee and An Empirical Study on Privacy and Fairness~\citep{lyu2020differentially} 
& Embedding Representations 
& No 
& No 
& Differential Privacy \\

DP-Forward: Fine-tuning and Inference on Language Models with Differential Privacy in Forward Pass~\citep{dp-forward}  
& Embedding Representations
& No 
& No 
& Differential Privacy \\

Flocks of Stochastic Parrots: Differentially Private Prompt Learning for Large Language Models~\citep{duan2023flocks} 
& In-Context Demonstrations 
& No 
& No 
& Differential Privacy \\

Privacy-Preserving In-Context Learning for Large Language Models~\citep{wuprivacy} 
& In-Context Demonstrations 
& No 
& No 
& Differential Privacy \\

Privacy-Preserving In-Context Learning with Differentially Private Few-Shot Generation~\citep{tangprivacy} 
& In-Context Demonstrations 
& No 
& No 
& Differential Privacy \\

Differential Privacy for Text Analytics via Natural Text Sanitization~\citep{yue2021differential} 
& Input Prompt 
& No 
& No 
& Differential Privacy \\

Privacy- and Utility-Preserving Textual Analysis via Calibrated Multivariate Perturbations~\citep{feyisetan2020privacy} 
& Input Prompt 
& No 
& No 
& Differential Privacy \\

DP-OPT: Make Large Language Model Your Privacy-Preserving Prompt Engineer~\citep{hong2024dpoptmakelargelanguage} 
& Prompt Tuning Data 
& No 
& No 
& Differential Privacy \\

DP-TabICL: In-Context Learning with Differentially Private Tabular Data~\citep{carey2024dp} 
& In-Context Demonstrations 
& Tabular Classification 
& No 
& Differential Privacy \\

Text Embeddings Reveal (Almost) As Much As Text~\citep{morris2023text} 
& Embedding Representations 
& Clinical Notes 
& Yes 
& Differential Privacy \\

InferDPT: Privacy-preserving Inference for Closed-box Large Language Models~\citep{tong2025inferdpt} 
& Input Prompt 
& No 
& Yes 
& Differential Privacy \\

Differentially Private In-Context Learning~\citep{panda2023differentially} 
& Input Prompt 
& Summarization 
& Yes 
& Differential Privacy \\

Sentence Embedding Leaks More Information than You Expect: Generative Embedding Inversion Attack to Recover the Whole Sentence~\citep{li2023sentence} 
& Embedding Representations 
& No 
& No 
& N/A \\

Does Prompt-Tuning Language Model Ensure Privacy?~\citep{xie2023does} 
& Prompt Tuning Data 
& Email Autoreply 
& Yes 
& N/A \\

Privacy Risks of General-Purpose Language Models~\citep{pan2020privacy} 
& Embedding Representations 
& No 
& Yes 
& N/A \\

On the Privacy Risk of In-Context Learning~\citep{duan2024privacy} 
& In-Context Demonstrations 
& No 
& Yes 
& N/A \\

Information Leakage in Embedding Models~\citep{song2020information}
& Embedding Representations 
& No 
& No 
& Sanitization/Obfuscation \\

TextObfuscator: Making Pre-trained Language Model a Privacy Protector via Obfuscating Word Representations~\citep{zhou-etal-2023-textobfuscator} 
& Embedding Representations 
& No 
& No 
& Sanitization/Obfuscation \\

Protecting User Privacy in Remote Conversational Systems: A Privacy-Preserving Framework Based on Text Sanitization~\citep{kan2023protecting} 
& Personally Identifiable Information (PIIs) 
& No 
& No 
& Sanitization/Obfuscation \\

Beyond Memorization: Violating Privacy via Inference with Large Language Models~\citep{staab2024beyond} 
& User Attributes 
& No 
& No 
& Sanitization/Obfuscation \\

LatticeGen: A Cooperative Framework Which Hides Generated Text in a Lattice for Privacy-Aware Generation on Cloud~\citep{zhang2023latticegen} 
& Output Generation 
& No 
& Yes 
& Sanitization/Obfuscation \\

Privacy-Preserving Language Model Inference with Instance Obfuscation~\citep{yao2024privacy} 
& Output Generation 
& No 
& Yes 
& Sanitization/Obfuscation \\

Hide and Seek (HaS): A Lightweight Framework for Prompt Privacy Protection~\citep{hideandseek} 
& Personally Identifiable Information (PIIs) 
& No 
& Yes 
& Sanitization/Obfuscation \\

EmojiPrompt: Generative Prompt Obfuscation for Privacy-Preserving Communication with Cloud-based LLMs~\citep{lin2025emojiprompt} 
& Input Prompt 
& Recommendation 
& Yes 
& Sanitization/Obfuscation \\

\bottomrule
\end{tabular}
\caption{Annotations for all 31 papers.}
\label{tab:drift_academic}
\end{table*}
\begin{table*}[t]
\centering
\small
\begin{tabular}{p{4.5cm} p{3.5cm} p{2.5cm} p{2.8cm} p{3cm}}
\toprule
\textbf{Title} & \textbf{Use Case} & \textbf{Price} & \textbf{Technique Category} \\
\midrule

\href{https://github.com/janhq/jan}{Jan} & General Purpose & Free & Local Deployment \\

\href{https://github.com/ggml-org/llama.cpp}{llama.cpp} & General Purpose & Free & Local Deployment \\

\href{https://github.com/mudler/LocalAI}{LocalAI} & General Purpose & Free & Local Deployment \\

\href{https://github.com/ollama/ollama}{Ollama} & General Purpose & Free & Local Deployment \\

\href{https://github.com/sgasser/pasteguard}{PasteGuard} & General Purpose & Free & Sanitization/Obfuscation \\

\href{https://github.com/Michael-A-Kuykendall/shimmy}{Shimmy} & General Purpose & Free & Local Deployment \\

\href{https://tinfoil.sh}{Tinfoil} & General Purpose & Paid & Cryptography \\

\href{https://duck.ai}{Duck.ai} & General Purpose & Freemium & Sanitization/Obfuscation \\

\href{https://phala.com}{Phala} & General Purpose & Paid & Cryptography \\

\href{https://github.com/protectai/llm-guard}{LLMGuard} & General Purpose & Free & Sanitization/Obfuscation \\

\href{https://chromewebstore.google.com/detail/privacygpt/iobeegngilbjemenccdplkjndkkpplbe?pli=1}{PrivacyGPT} & General Purpose & Free & Sanitization/Obfuscation \\

\bottomrule
\end{tabular}
\caption{Annotations for all 11 tools.}
\label{tab:drift_tools}
\end{table*}

\end{document}